\documentclass{article}
\usepackage[utf8]{inputenc}

\usepackage{times}
\usepackage{microtype}
\usepackage{graphicx}

\usepackage{subcaption}
\usepackage{booktabs} 
\usepackage{algorithmic}
\usepackage{multirow}
\usepackage{amsmath}
\usepackage{comment}
\usepackage{caption}
\usepackage[T1]{fontenc}
\usepackage{authblk}
\usepackage{bm}

\def \x {\mathbf{x}}
\def \w {\mathbf{w}}
\def \R {\mathcal{R}}

\title{Defend Deep Neural Networks Against Adversarial Examples via Fixed and Dynamic Quantized Activation Functions}

\author[1]{Adnan Siraj Rakin}
\author[2]{Jinfeng Yi}
\author[3]{Boqing Gong}
\author[1]{Deliang Fan\thanks{Corresponding Author: dfan@ucf.edu}}
\affil[1]{Department of Computer Engineering, University of Central Florida}
\affil[2]{JD AI Research, Beijing, China}
\affil[3]{Tencent AI Lab, Bellevue, WA, USA}

\date{}

\usepackage{hyperref}

\usepackage{import}
\usepackage{tikz}
\usetikzlibrary{positioning}

\usepackage[capitalise]{cleveref}

\usepackage{geometry}

\begin{document}

\maketitle

\begin{abstract}
Recent studies have shown that deep neural networks (DNNs) are vulnerable to adversarial attacks. To this end, many defense approaches that attempt to improve the robustness of DNNs have been proposed. In a separate and yet related area, recent works have explored to quantize neural network weights and activation functions into low bit-width to compress model size and reduce computational complexity. In this work, we find that these two different tracks, namely the pursuit of network compactness and robustness, can be merged into one and give rise to  networks of both advantages. To the best of our knowledge, this is the first work that uses quantization of activation functions to defend against adversarial examples. We also propose to train robust neural networks by using adaptive  quantization techniques for the activation functions. Our proposed Dynamic Quantized Activation (DQA) is verified through a wide range of experiments with the MNIST and CIFAR-10 datasets under different white-box attack methods, including FGSM, PGD, and C\&W attacks. Furthermore, Zeroth Order Optimization and substitute model based black-box attacks are also considered in this work. The experimental results clearly show that the robustness of DNNs could be greatly improved using the proposed DQA. 


\end{abstract}

\section{Introduction}
\begin{figure}[ht]
  \centering
   \includegraphics[width=0.5\textwidth,height=0.17\textwidth]{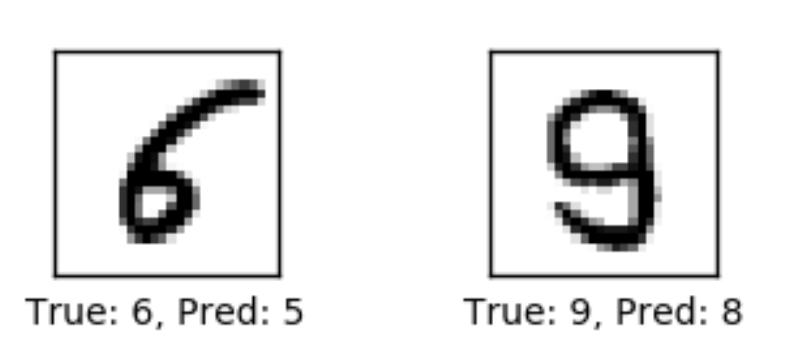}
   \caption{Adversarial attack miss-classifying MNIST digits}
   \label{mnisti}
\end{figure}

Deep Neural Networks (DNNs) have achieved great success in various tasks, including but not limited to image classification~\cite{krizhevsky2012imagenet},  speech recognition \cite{hinton2012deep}, machine translation~\cite{bahdanau2014neural}, and autonomous driving~\cite{chen2015deepdriving}. Despite the remarkable progress, recent studies~\cite{szegedy2013intriguing,goodfellow2014explaining,carlini2017towards} have shown that DNNs are 
vulnerable to adversarial examples. In image classification, an adversarial example is a carefully crafted image that is visually imperceptible to the original image but can cause DNN model to misclassify as shown in figure \ref{mnisti}. In addition to image classification, attacks to other DNN-related tasks have also been actively investigated, such as visual QA~\cite{xu2017can}, image captioning~\cite{chen2017show},  semantic segmentation~\cite{metzen2017universal}, machine translation~\cite{cheng2018seq2sick}, speech recognition~\cite{ carlini2018audio}, and medical prediction~\cite{sun2018identify}.  

There is a cohort of works on generating adversarial attacks and developing corresponding defense methods. The adversarial attacks can be grouped into two major categories: (i) \textit{white-box attack}~\cite{szegedy2013intriguing,carlini2017towards}, where the adversary has full access to the network architecture and parameters, and (ii) \textit{black-box attack}~\cite{liu2016delving,papernot2016transferability,chen2017zoo}, where the adversary can access the input and output of a DNN but not its internal configurations. Many attack algorithms have been proposed to generate adversarial examples~\cite{szegedy2013intriguing,kurakin2016adversarial,kos2017delving,papernot2016limitations,moosavi2016deepfool,athalye2018obfuscated,chen2017ead,carlini2017towards}. Among them, the fast gradient sign method (FGSM)~\cite{goodfellow2014explaining} is one of the pioneering and most popular white-box attack algorithms. It uses the sign of gradients with respect to the input to generate adversarial examples and is by far one of the most efficient attack algorithms. The projected gradient descent (PGD) is among the most powerful white-box attacks till date~\cite{madry2017towards}. Besides, Carlini \& Wagner (C\&W)~\cite{carlini2017towards} attack is another powerful attack that can achieve nearly 100\% attack success rate. In this work, we employ all these attack algorithms to evaluate the robustness of our proposed defense. We also conduct popular black box attacks to even further establish our claim of improved robustness \cite{chen2017zoo,papernot2017practical}. To achieve the same objective previously many defense methods have been proposed to defend against adversarial examples~\cite{gu2014towards,papernot2016distillation,papernot2016towards,xu2017feature}. Some of them work well against early attack methods (e.g., FGSM~\cite{goodfellow2014explaining} and Jacobian saliency map attack (JSMA)~\cite{papernot2016limitations}), but fail to deliver strong robustness against more powerful attacks~\cite{madry2017towards,carlini2017towards,athalye2018obfuscated}.


Meanwhile, from another perspective, DNNs with increasing depth, larger model size, and more complex building blocks have been proposed to achieve better performance and perform more sophisticated tasks. For instance, Inception-ResNet~\cite{szegedy2017inception} and DenseNet~\cite{huang2017densely} have 96 and 161 layers, and 56.1 million and 29 million parameters, respectively. The large sizes prohibit them from deploying in resource intensive systems such as mobile phones. Fortunately, DNN models usually contain significant redundancy, and the state-of-the-art accuracy can also be achieved after model compression~\cite{han2015deep}. 
As one of the most popular model compression techniques, weight and activation quantization is widely explored to significantly shrink the model size and reduce the computational cost~\cite{courbariaux2015binaryconnect,rastegari2016xnor,zhou2016dorefa}. BinaryConnect \cite{courbariaux2015binaryconnect} introduces gradient clipping and is the first binary CNN algorithm that yields close to the state-of-the-art accuracy on CIFAR-10. After that, both BWN in \cite{rastegari2016xnor} and DoreFa-Net \cite{zhou2016dorefa} show better or close validation accuracy on ImageNet dataset. We note that although weight quantization is more widely-used in reducing model size, quantization of activation functions can also make network more compact since the input bit-width of next layer is determined by the quantization level of activation functions. In this work, we mainly consider activation quantization as an effective defense methodology against adversarial attacks. 

In this work, we show that the tracks of pursuing network compactness and robustness are likely to merge, although they seem to be independent. Specifically, we propose an activation quantization technique called Dynamic Quantizated Activation (DQA), which treats thresholds of activation quantization function as another tuning parameter set to improve the robustness of DNN in adversarial training. In this method, such threshold parameters are adapted in adversarial training process and play an important role in improving network robustness against adversarial attacks. In this work, we first test fixed activation quantization method under a wide range of attacks. It shows empirically that, by quantizing the output at each activation layer, the effect of adversarial noises can be greatly diminished. To the best of our knowledge, this is the very first work that proposes to use quantized activation function to defend against adversarial examples. Then, we show that the proposed DQA can further supplement this claim by including more learnable parameters during adversarial training process. By doing so, the compactness and robustness of neural networks are simultaneously achieved.

\section{Related Works}

Recent development of some of the strongest \cite{madry2017towards,carlini2017towards,athalye2018obfuscated,chen2017zoo} attacks have exposed the underlying vulnerability of DNN more than ever. As a result, a series of work have been conducted towards the development of robust DNN \cite{samangouei2018defense,papernot2016transferability,madry2017towards,qian2018l2,santhanam2018defending,xie2017mitigating,guo2017countering,tramer2017ensemble,raghunathan2018certified,dhillon2018stochastic,prakash2018deflecting}. Among them, training the model with adversarial examples gives the initial breakthrough towards universal robustness of DNN \cite{madry2017towards}. Most of later works have followed this path to supplement their defense with adversarial training. Other defenses have focused on transforming the input to DNN \cite{buckman2018thermometer,guo2017countering}. However, input transformation seems to suffer from obfuscated gradient issue \cite{athalye2018obfuscated}. Recent development of Generative Adversarial Network (GAN) have showed promising signs for GAN networks to work as a potential defense model \cite{santhanam2018defending,samangouei2018defense}. One of the most recent defense that is similar to our work is L2 non-expansive neural networks \cite{qian2018l2}. The principle of this defense is to suppress the noise inside DNN and stop it from becoming large as it goes through the network. Our idea behind introducing quantization is similar. However, due to the hard constraints of L2 non-expansive defense, it suffers from poor clean data accuracy. On the other hand, previous works have shown quantization of activation does not hamper the clean data accuracy much \cite{zhou2016dorefa} compared to L2 non-expansive defense.

\section{Fixed and Dynamic Activation Quantization}

In this section, we first describe the fixed quantization technique and its role in working as a defense against adversarial examples. Then, we present the working principle of our proposed Dynamic Quantized Activation as an effective methodology to improve the robustness of DNN against adversarial examples.

\subsection{Fixed Quantization}

\paragraph{Fixed Method:}
We start from a fixed activation quantization method that shares some similarities with Dorefa-Net \cite{zhou2016dorefa}. 
The inputs to quantization function are first passed through a tanh function that maps the input $x$ to range (-1, 1):
\begin{eqnarray}
y=tanh(x)&=&\frac{sinh(x)}{cosh(x)}\\\nonumber
&=&\frac{e^x-e^{-x}}{e^x+e^{-x}}
\end{eqnarray}

\noindent We then shift the output $y$ to range of $(0,\ 1)$ using the following function:

\begin{equation}
\label{eq:f}
y'=f(y)={\frac{y}{2\max(|y|)}+\frac{1}{2}}.
\end{equation}

\noindent Where, $max(|y|)$ denotes the maximum values in the y matrix without sign. Then an $n$-bit quantization is achieved by:
\begin{equation}
\label{eq:2}
y_n=f(y')={2\times[\frac{1}{2^n-1}\times round[(2^n-1)\times(y')]]-1},
\end{equation}

\noindent In this way, the output tensors will only contain $2^n$ discrete levels in the range of -1 to +1. For example, when $n=2$, the outputs will be quantized to 4 discrete levels -1, -0.33, 0.33,and 1.

\paragraph{A Simple Explanation}
Although the qunatization method is very simple, they work surprisingly well in improving the robustness of neural networks as will be discussed in the experiment section. Here, we first provide a simple explanation of this phenomenon. 

In deep neural networks, a small change in the inputs will usually cause significant changes in the outputs. Existing works~\cite{qian2018l2,goodfellow2014explaining} have shown that this is one of the main reasons why adversarial examples exist. To see this, let's consider a simple neuron 
\begin{equation}
y=f(\x^\top\w),   
\end{equation}

where $\x\in \R^d$ is the input signal, $\w=[w_1,\cdots,w_d]^\top$ is the neuron weights, and $f(\circ)$ denotes the widely-used rectified linear unit (ReLU) activation function
\begin{equation}
{ReLU(z)}=
\begin{cases}
    z,& \text{if } z \geq 0\\
    0,          & \text{otherwise}.
\end{cases}
\end{equation}

\noindent When adding a small perturbation $\boldsymbol{\epsilon}\in \R^d$ to the input vector $\x$, the neuron output becomes 
\begin{equation}
\tilde y=f[(\x+\boldsymbol{\epsilon})^\top\w]=f[\x^\top\w+\boldsymbol{\epsilon}^\top\w]  
\end{equation}

When each element of $\x^\top\w+\boldsymbol{\epsilon}^\top\w$ is larger than 0, this makes the neuron output to grow by as large as $\boldsymbol{\epsilon}^\top\w$. If the average magnitude of the weight vector element is $a$ and $\|\boldsymbol{\epsilon}\|_\infty= r$, then the neuron output will grow by $r\times d\times a$, which is non negligible when the input dimensionality is large. Since a deep neural network usually contains a large number of neurons and layers, a small change in the inputs usually causes a very significant change in the outputs. 

Indeed, this observation also  motivates some recent defense techniques. For example, \cite{qian2018l2} proposes an $\ell_2$-nonexpansive neural network that enforces a unit amount of change in the inputs can cause at most a unit amount of change in the outputs. However, this non expansive constraint is so strong that may hurt the expressibility and learnability of neural network~\cite{lin2017does}. As a result, its performance on clean data is largely sacrificed. In contrast, our method addresses the exploding output problem by using quantized activation functions, a technique widely-used in model compression. By doing so, the proposed method can simultaneously achieve two goals: (i) improving the robustness of neural networks; (ii) making neural networks more compact. 

\subsection{Dynamic Activation Quantization}

\paragraph{Adversarial Training:}

Given a set of training images and lables $(\x,\ y)$ sampled from the distribution $L$ and a loss function $J$, the adversarial training method aims to learn parameters $\gamma$ to minimize the risk $R_{(\x,\ y)}\sim_L[J(\gamma,\ \x,\ y)]$. Typically,  $\gamma$ in a DNN consists of weights (W), biases (b) and other parameters like batch normalization layer parameters (z).

The first step in adversarial training is to choose an attack model to generate adversary examples. In this work, we employ the PGD attack since it can generate universal adversary examples among the first order approaches \cite{madry2017towards}. Using this model, for each image $\x$, the adversary example $\x$+ $\epsilon$ within the bounded $\ell_\infty$ norm is generated. 
Similar to \cite{kurakin2016adversarial, madry2017towards,wald1945statistical}, given the generated adversarial examples, we aim to minimize the following empirical risk to improve the model robustness
\begin{eqnarray}
\label{eqn:5}
{\min}\ R_{(\x,y)}\sim_L[\max \ J(\gamma,\x+\epsilon,y)].
\end{eqnarray}
The optimal solution of the above min-max problem is achieved by tuning parameters such as weights and biases. According to our previous analysis, quantized activation functions may also improve the network robustness. It motivates us to integrate the quantized activation functions into adversarial training, and treat the thresholds between different quantization levels as another set of tunable parameters to improve network robustness.


\paragraph{Dynamic Quantized Activation:}
In our previous method, the thresholds between different quantization levels are fixed and uniformly distributed. Here we propose a dynamic activation quantization method in which the thresholds for different discrete levels are tunable parameters in adversarial training. 
For n-bit quantized activation function, it has $2^n$ output discrete levels and $2^n-1$ tunable thresholds. Let $m=2^n$, then the quantization will have m-1 threshold values $t_1,\ t_2,\ ..,\ t_{m/2},\ ...,\ t_{m-2}$ and $ t_{m-1}$. Then, we make $m$ level quantization function as:
\begin{eqnarray}
\label{eqn:6}
f(x)&=&0.5\times[sgn(x-t_{m-1})+ \nonumber\\
&&\sum_{i=m-1}^{m/2+1} t_i(sgn(t_i-x)+sgn(x-t_{i-1}))\nonumber\\
&&+\sum_{i=m/2}^{2} t_{i-1}(sgn(t_i-x)\nonumber
+sgn(x-t_{i-1}))-\nonumber\\
&&sgn(t_1-x)], 
\end{eqnarray}
where $sgn$ denotes the sign function. For example, when $n=1$ and $2$, the $1$-bit and $2$-bit dynamic quantizations are: 
\begin{align*}
f(x)=0.5\times[sgn(x-t_1)-sgn(t_1-x)]
\end{align*}
and
\begin{equation}
{f(x)}=
\begin{cases}
    -1,& \text{if } x \ < \ t_1\\
    t_1,& \text{if } t_1\ < \ x \ < \ t_2\\
    t_3,& \text{if } t_2\ <\ x \ < \ t_3\\ 
    +1,& \text{if } x \ > \ t_3,\\ 
\end{cases}
\end{equation}
respectively. Therefore, equation (\ref{eqn:6}) adds a new set of learnable parameters $T$ := [$t_1,t_2,...,t_{m-1}$]. Since they can be learned independently w.r.t the existing DNN parameters $\gamma$, the objective function (5) then becomes:

\begin{equation}
{\min}\ R_{(\x,y)}\sim_L[\max \ J([\gamma,T],\x+\epsilon,y)],
\end{equation}

which is more flexible and desirable than the previous approach that can only tune DNN parameters $\gamma$.

\begin{figure}[ht]
  \centering
   \includegraphics[width=0.5\textwidth]{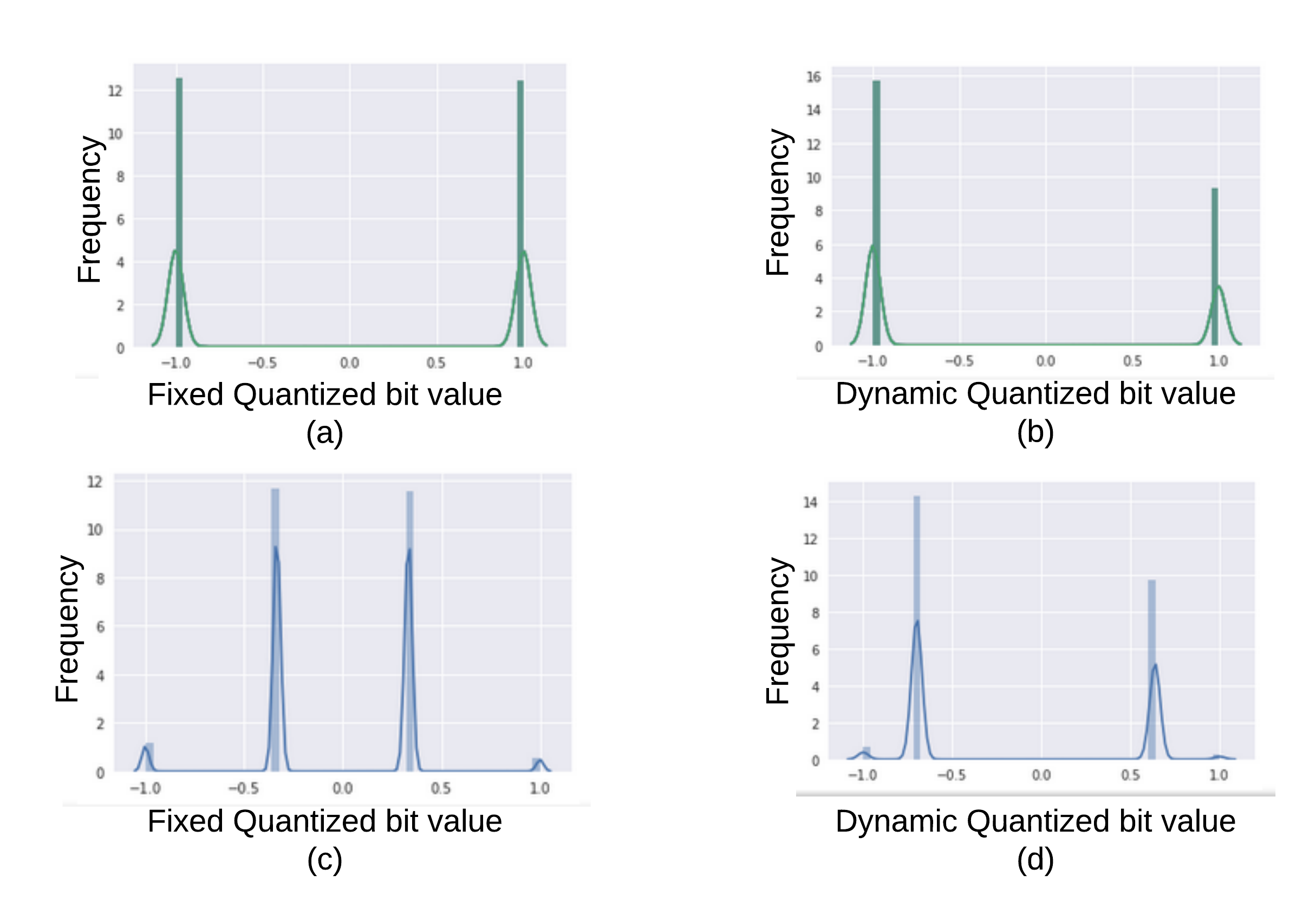}
   \caption{Activation layer distributiion: a)1-bit fixed threshold b)1-bit dynamic threshold c)2-bit fixed threshold and d) 2-bit dynamic threshold}
   \label{dis}
\end{figure}

To illustrate the effect of dynamic activation quantization, as shown in figure \ref{dis}, we plot the data distribution after the first activation layer of a sample network for one random input patch in CIFAR10 dataset. For both 1-bit and 2-bit activation quantization, we show the distribution first using fixed quantized activation. Then we replace the fixed activation with dynamically trained activation to indicate the difference of data distribution. It is clear that fixed quantized levels are uniformly distributed between -1 to +1. On the other hand, in dynamic activation quantization method, discrete levels are adjusted and the frequency at each quantized levels are not even as well. Such adjusted distribution reflects the quantization function's involvement in the adversarial training and its tuning to the network performance. Its efficacy in improving network robustness will be verified in later experimental section.

\section{Experimental Setup}

\paragraph{Datasets and networks.}
We first test LeNet~\cite{lecun2015lenet} on the MNIST dataset~\cite{lecun1995convolutional}. MNIST is a set of handwritten digit images with 60,000 training examples and 10,000 testing examples. Besides, we test ResNet-18~\cite{he2016deep} on the CIFAR-10 dataset~\cite{krizhevsky2010cifar}. CIFAR-10 contains 60,000 RGB images of size 32x32. Following the standard practice, we use 50,000 examples for training and the remaining 10,000 for testing. The images are drawn evenly from 10 classes. Since we do not have a validation set for our experiments we run the simulations five times for each case to report the average of them. 

\paragraph{Attack methods.}
To evaluate the robustness of our method, we employ multiple powerful white-box attack methods, including the projected gradient descent (PGD) attack~\cite{madry2017towards}, the fast gradient sign method (FGSM)~\cite{goodfellow2014explaining}, and the Carlini \& Wagner (C\&W) attack~\cite{carlini2017towards}. We also conduct state-of-the-art black box attacks  in a seperate section as well \cite{papernot2017practical,chen2017zoo} to test our defenses robustness. The parameters of all the attack algorithms are tuned to give the best attack performance.

\paragraph{Defense techniques and baselines.}
We conduct experiments for our defense including adversarial training. Because the very essence of DQA working as a defense requires the model to be trained with adversarial training. In adversarial training, the adversarial examples are generated by the PGD attack algorithm. Before mixing adversarial examples in the training, we train the model with clean data for some initial epoch of training. 

We compare our defense result with the original PGD trained defense model \cite{madry2017towards} that uses ReLu as the full-precision activation function. Later, we also show how our defense compares with other state-of-the-art defenses in defending the popular attacks.

\section{White Box Attack}
We test our idea of introducing quantized activation as a defense against adversarial attack using Lenet and Resnet-18 architecture for MNIST and CIFAR10, respectively. Adversarial training is incorporated in both fixed and dynamic activation quantization method for a fair comparison.  

\begin{table}[ht]
\begin{tabular}{|l|l|l|l|l|l|l|}
\hline
\multirow{2}{*}{Activation Bits} & \multicolumn{2}{l|}{Clean} & \multicolumn{2}{l|}{PGD} & \multicolumn{2}{l|}{FGSM} \\ \cline{2-7} 
 & Fixed & Dynamic & Fixed & Dynamic & Fixed & Dynamic \\ \hline
1-bit & 98.52 \% & 98.53\% & 98.47\% & 98.53\% & 98.48\% & 97.44\% \\ \hline
2-bit & 98.65\% & 98.80\% & 98.46\% & 98.75\% & 98.59\% & 98.75\% \\ \hline
3-bit & 99.05\% & 99.01\% & 98.37\% & 98.51\% & 98.7\% & 97.15\% \\ \hline
Full-Precision & \multicolumn{2}{c|}{99.24\%} & \multicolumn{2}{c|}{94.04\%} & \multicolumn{2}{c|}{97.1\%} \\ \hline
\end{tabular}
\centering
\caption{Summary of DQA's preformance as a defense for MNIST}
\label{tab:2}
\end{table}

\subsection{Results on MNIST Dataset}
We run two different sets of simulation for fixed and dynamic activation quantiztion, where each consists of different bit-widths of quantization levels. First, we obtain the results using ReLu activation function reported as 'full-precision' activation in table \ref{tab:2}. Later we report the clean data and under-attack accuracy for 1-, 2-, 3-bit fixed and dynamic activation quantization. The summary of simulation results under both PGD and FGSM attack for MNIST dataset is shown in Table \ref{tab:2}. PGD based adversarial training is used as a base defense method that achieves about 94\% accuracy under PGD attack and 97.1\% accuracy under FGSM attack. It can be seen that, after incorporating activation quantization, the accuracies under both attacks are easily increased to close to 98\%. For example, 1-bit fixed activation quantization could improve the accuracy from 94.04\% to 98.47\% under PGD attack. Meanwhile, slight improvements are observed when dynamic activation functions are used for all the cases. Both methods could improve $\sim$4\% accuracy compared to baseline (i.e. full-precision) under PGD attack. But dynamic quantization activation function results in a stronger defense for PGD than that for FGSM attack for 1-bit and 3-bit activation. However, by using 2-bit activation we could get the best defense accuracy of 98.75 \%  for FGSM attack.

\subsection{Results on CIFAR10}
We also test our defense method using Resnet-18 architecture for cifar10 dataset. As a baseline case (i.e. full-precision), we similarly employ ReLu activaiton function. Then, both fixed and dynamic activation quantization methods are simulated and tabulated in table \ref{my-label}.

\begin{table}[ht]
\begin{tabular}{|l|l|l|l|l|l|l|}
\hline
\multirow{2}{*}{Activation Bits} & \multicolumn{2}{l|}{Clean} & \multicolumn{2}{l|}{PGD} & \multicolumn{2}{l|}{FGSM} \\ \cline{2-7} 
 & Fixed & Dynamic & Fixed & Dynamic & Fixed & Dynamic \\ \hline
1-bit & 81.43 \% & 83.16\% & 72.4\% & 77.62\% & 72.9\% & 76.09\% \\ \hline
2-bit & 82.33\% & 85.76\% & 70.51\% & 79.83\% & 71.8\% & 76.37\% \\ \hline
3-bit & 84.14\% & 84.17\% & 63.23\% & 75.01\% & 73.18\% & 74.51\% \\ \hline
Full-Precision & \multicolumn{2}{c|}{87.59\%} & \multicolumn{2}{c|}{48.70\%} & \multicolumn{2}{c|}{58.16\%} \\ \hline
\end{tabular}
\centering
\caption{Summary of DQA's preformance as a defense for CIFAR10}
\label{my-label}
\end{table}

First, similar as many other defense methods \cite{madry2017towards,qian2018l2}, activation quantization method will slightly degrade clean data accuracy. For example, the clean data accuracy degrades from 87.59\% to 81.43\% when fixed binary activation function is used. But, it is interesting to also observe that such accuracy degradation effect could be mitigated when we introduce dynamic activation quantization as shown in table \ref{my-label}. For example, it improves binary activation accuracy from 81.43\% to 83.16\%. Moreover, DQA could get as high as 85.76 \% clean data accuracy using 2-bit dynamic quantized activation function.

Second, for fixed activation quantization function, the defense against adversarial examples increases with decreasing quantization bit-width. For example, under strong PGD attack, CIFAR 10 accuracy increases up to 72.4\% for 1-bit activation function, from 48.7\% for full precision activation function. The defense decreases significantly as we increase the activation  quantization bit width. For example, a 3-bit activation quantization leads accuracy to drop down to 63.23\% under PGD attack. However, it is still 14.53\% better than full precision activation function. A similar trend was also observed for FGSM attack as well.

Third, for dynamic activation quantization function, we observe a twofold improvement. First, dynamic activation recovers the clean data accuracy to some extent. Second, as explained earlier, adding dynamic quantization thresholds to the loss function works in favor of model robustness. Our experimental results show accuracy improvement for all the low bit width activation function under PGD and FGSM attacks, compared to fixed bit quantization. Employing DQA method, we could push the accuracy up to 79.83\% using 2-bit dynamic activation function across all layers. Again, for CIFAR10, we observe FGSM attack becomes stronger compared to PGD which is similar for MNIST as well. However, around 4\% higher accuracy is observed with dynamic activation quantization function for both 1-bit and 2-bit cases when attacked using FGSM.

\begin{figure}[ht]
  \centering
   \includegraphics[width=0.5\textwidth,height=0.45\textwidth]{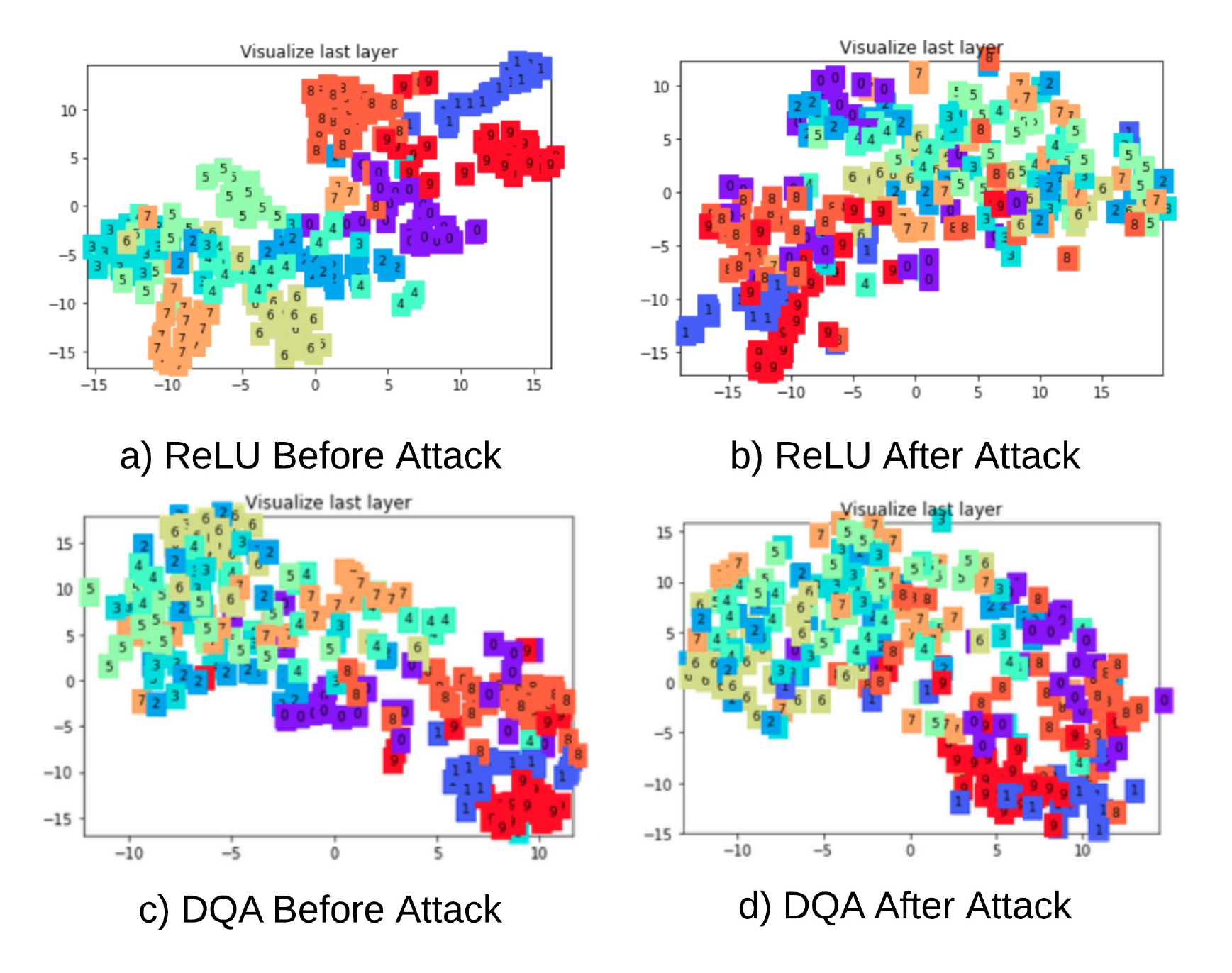}
   \caption{tSNE visualization before and after attack using both Relu and DQA activation}
   \label{tsne}
\end{figure}

In order to better illustrate how DQA could help improve DNN robustness, we employ t-SNE to visualize input data of last fully connected layer for CIFAR10 dataset as shown in figure \ref{tsne}. t-SNE is a popular data visualization technique to get more insight to high dimensional data by reducing the dimensionality \cite{maaten2008visualizing}. Typical shape and distance of a t-SNE plot does not carry much significance. Rather, for classification tasks, the data that belongs to the same class should create a cluster of points close to each other. A network performing well on the classification task should create a cluster without much scattering. We plot the same data using both ReLU and DQA function before attacking in figure \ref{tsne}(a)\&(c) and after attacking in figure \ref{tsne}(b)\&(d). For both data in figure \ref{tsne}(a)\&(c), it is clear that the data points belong to the same group stay in a close cluster before attack. However, under adversarial attack, these data become much more scattered in traditional ReLU activation function compared to our proposed dynamic quantized activation function.

\paragraph{C \& W Attack}

We also conduct experiments of our proposed defense methods under C \& W attack \cite{carlini2017towards}. It is an effective optimization-based attack model. The amount of perturbation to an input image is generated through $L_2$ norm by solving an optimization problem based on an objective function. The purpose of such objective function is to find adversary that would miss-classify the image. A higher value of $L_2$ norm indicates a more robust network or potential failure of the attack. 
We measure the averaged $L_2$ norm required to successfully attack each cases for Cifar 10 dataset under C \& W attack in this subsection.

\begin{table}[ht]
\centering
\caption{C \& W attack summary:}
\label{tab:5}
\begin{tabular}{|c|c|c|}
\hline
\multicolumn{1}{|l|}{Activation} & \multicolumn{1}{l|}{\begin{tabular}[c]{@{}l@{}}Fixed quantization\\ Without PGD training\\ (L2 norm)\end{tabular}} & \multicolumn{1}{l|}{\begin{tabular}[c]{@{}l@{}}DQA \\ With PGD training\\Attack Succ. Rate\end{tabular}} \\ \hline
32-bit & 0.12 & 0 \\ \hline
3-bit & 0.17 & 0 \\ \hline
2-bit & 0.26 & 0 \\ \hline
1-bit & 0.55 & 0 \\ \hline
\end{tabular}
\end{table}

Table \ref{tab:5} shows the $L_2$ norm required for C \& W attack to get 100 percent success in different cases in the first column. Again, for C \& W attack, we observe the same pattern as we observed before for both PGD and FGSM attacks. Average $L_2$ norm required for 100 percent successfull attack keeps decreasing with increasing activation quantization bit width. We also run the experiments of dynamic activation quantization under C \& W attack using adversarial training. But our proposed DQA successfully defends C \& W attack as shown in the second column. The C \& W attack fails to miss-classify the input with reasonable $L_2$ norm within a certain amount of time when supplemented by PGD training. As explained in \cite{madry2017towards}, defenses trained with $l_\infty$ bounded universal attack tends to perform well against $l_2$ bounded attacks, resulting in high $l_2$ norm which are eventually detectable even by human eyes.

\subsection{Comparison to other state-of-the-art defense methodologies}

In this subsection, we compare our proposed defense performance with several recent popular defense methodologies. One of the most successful defense is madry's defense \cite{madry2017towards} that uses PGD adversarial training as the main form of defense. Such adversarial training method has become the standard for most recent defense techniques to improve DNN robustness. For example, one such defense is thermometer encoding that performs encoding of input data as a defense \cite{buckman2018thermometer}. However, their proposed method has gradient scattering issue \cite{athalye2018obfuscated}. In another work called defense GAN, it uses GAN discriminator and generator to counter the adversarial examples \cite{samangouei2018defense}. Cowboy is another GAN based defense that uses the scores from a GAN's discriminator to separate adversary from clean examples \cite{santhanam2018defending}. Note that, all of the defenses here include PGD based adversarial training except the Cowboy defense. Here, we report accuracy under PGD and FGSM attack mdoels for different defenses in table \ref{tb:cmp}, where the values of $\epsilon$ are 0.3 and 0.031, respectively, for MNIST and CIFAR10 datasets. Meanwhile, all of the defense accuracy corresponding to PGD attack employ attack steps of 40 and 7 for MNIST and CIFAR10, respectively.

\begin{table}[ht]
\centering
\caption{Our defense performance compared to the state of art defenses}
\label{tb:cmp}
\begin{tabular}{|l|l|l|l|l|}
\hline
Defense & Arch. & \begin{tabular}[c]{@{}l@{}}Clean\\ \%\end{tabular} & \begin{tabular}[c]{@{}l@{}}(PGD)\\ \%\end{tabular} & \begin{tabular}[c]{@{}l@{}}FGSM\\ \%\end{tabular} \\ \hline
\multicolumn{5}{|l|}{MNIST} \\ \hline
Madry \cite{madry2017towards} & Vanilla & 98.80 & 93.2 & 95.6 \\ \hline
T. E. \cite{buckman2018thermometer}& Vanilla & 99.03 & 94.02 & 95.82 \\ \hline
\begin{tabular}[c]{@{}l@{}}Defense\\ GAN \cite{samangouei2018defense}\end{tabular} & GAN & 99.20 & \_\_\_ & 98.43 \\ \hline
Cowboy\cite{santhanam2018defending} & GAN & 97 & 78 & 81 \\ \hline

\begin{tabular}[c]{@{}l@{}}DQA\\ (Ours)\end{tabular} & \begin{tabular}[c]{@{}l@{}}Lenet\\ (2-bit)\end{tabular} & 98.80 & 98.75 & 98.75 \\ \hline
\multicolumn{5}{|l|}{CIFAR10} \\ \hline
Madry\cite{madry2017towards} & \begin{tabular}[c]{@{}l@{}}Wide \\ Resnet\end{tabular} & 87.3 & 50.00 & 56.1 \\ \hline
T.E.\cite{buckman2018thermometer} & \begin{tabular}[c]{@{}l@{}}Wide \\ Resnet\end{tabular} & 89.88 & 79.16 & 80.96 \\ \hline
Cowboy\cite{santhanam2018defending} & GAN & 78 & 53 & 53 \\ \hline
\begin{tabular}[c]{@{}l@{}}DQA\\ (Ours)\end{tabular} & \begin{tabular}[c]{@{}l@{}}Resnet\\ 18 (2bit)\end{tabular} & 85.76 & 79.83 & 76.37 \\ \hline
\end{tabular}
\end{table}

Table \ref{tb:cmp} summarizes the comparison results of our proposed DQA defense methods over other recent popular defense methods. Since our methodology involves quantization of activation function, we expect a lower clean data accuracy especially when supplemented by adversarial training. However, our clean data accuracy of 99.01 \% and 85.76 \% for MNIST and CIFAR10 respectively are still within 0.5-2 \% of recent defenses. Our major success lies in achieving greatly improved accuracy in defending PGD attack. For both MNIST and CIFAR10, our reported accuracies are higher than that of all previous defenses reported here. As for FGSM attack, our method gives higher accuracy for MNIST than all of the reported attacks. In case of CIFAR, we could achieve better accuracy than PGD defense \cite{madry2017towards} and cowboy \cite{santhanam2018defending} as well. Another defense L2NNN \cite{qian2018l2} that uses almost similar theory as our defense of suppressing the input noise from increasing can actually get 91.7 \% and 22 \% accuracy for MNIST and CIFAR10 dataset, respectively, against l2 bounded adversarial attack without adversarial training. However, their clean data accuracy suffers even badly than ours reaching only 72 \% for CIFAR10 accuracy. The attempt to suppress the noise early in the network would result in hampering the clean data accuracy. Thus DQA's achievement of more than 85 \% accuracy for CIFAR with improved defense becomes even more significant.

\section{Black-Box Attack}
In this section, we also test our defense model against a variety of black box attacks. In black box setup, we assume that the attacker has no knowledge of the target DNN.
Even in a black box setup the attacker does not always need to train a substitute model, rather they can directly estimate the gradient of the target DNN model based on input and output scores \cite{chen2017zoo}.
In another form of black box attack, the attacker can always train a substitute model to attack the target model which we call the source \cite{papernot2017practical}.

\paragraph{ZOO Black Box Attack}
We also conduct zeroth order optimization(ZOO) attack on our defence \cite{chen2017zoo}. ZOO is a form of black box attack that does not require training a substitute model, rather it can approximate the gradient of the target DNN just based on input image and output scores. We use Zeroth Order Stochastic Coordinate
Descent with Coordinate-wise ADAM. Similar as the original paper, we test our defense on random 200 samples for untargetted attack to observe the effectiveness of the attack on proposed defense. We summarize the attack's success rate in table \ref{b:21} for CIFAR 10 dataset.

\begin{table}[ht]
\centering
\caption{CIFAR 10 accuracy for ZOO Attack}
\label{b:21}
\begin{tabular}{|c|l|l|l|}
\hline
Bit -Width & \multicolumn{1}{c|}{DNN} & \begin{tabular}[c]{@{}l@{}}Clean \\ Data\\ acc.\end{tabular} & \multicolumn{1}{c|}{\begin{tabular}[c]{@{}c@{}}Attack \\ Success \\ rate (\%)\end{tabular}} \\ \hline
32-bit & Res-18 & 87.59 & 81.75 \\ \hline
3-bit & Res-18 & 84.17 & 0 \\ \hline
2-bit & Res-18 & 85.76 & 0 \\ \hline
1-bit & Res-18 & 83.16 & 0 \\ \hline
\end{tabular}
\end{table}

It can be seen that our proposed dynamic quantization successfully defends against ZOO attack. Similar as the C \& W attack, ZOO- attack fails to approximate the gradient due to the intermediate quantization function. As a result it fails to attack the defence with any kind of success.

\paragraph{Black-Box Attack Using a Substitute Model:}
In this subsection, we conduct black box attack, we train a substitute model to perform the exact same classification task as the target. The approach is similar to the popular approach for black box attack \cite{papernot2017practical}. In our experiments, we separately train VGG16 and Alexnet models with full precision activation function to use it as source models. 

\begin{table*}[ht]
\centering
\caption{Black Box attack on Cifar 10 dataset using substitute model}
\label{tb}
\begin{tabular}{|l|l|l|l|l|}
\hline
Bit-width & \begin{tabular}[c]{@{}l@{}}Our Defense \\ Accuracy \\  No attack (\%)\end{tabular} & \begin{tabular}[c]{@{}l@{}}Substitute \\ Accuracy\\ No attack (\%)\end{tabular} & \begin{tabular}[c]{@{}l@{}}Substitute \\ Accuracy \\ Under \\ Attack (\%)\end{tabular} & \begin{tabular}[c]{@{}l@{}}Defense\\ Accuracy \\ Under \\ Attack(\%)\end{tabular} \\ \hline
\multicolumn{5}{|l|}{VGG16 as Substitute Model} \\ \hline
3-bit & 84.93 & 84.32 & 56.88 & 60.85 \\ \hline
2-bit & 85.06 & 85.06 & 56.62 & 60.46 \\ \hline
1-bit & 83.16 & 83.09 & 44.94 & 49.18 \\ \hline
\multicolumn{5}{|l|}{Alexnet as Substitute Model} \\ \hline
3-bit & 84.93 & 84.05 & 46.62 & 82.59 \\ \hline
2-bit & 85.06 & 85.03 & 52.42 & 82.11 \\ \hline
1-bit & 83.16 & 83.05 & 36.06 & 79.74 \\ \hline
\end{tabular}
\end{table*}

The summary of the results obtained using a substitute model is reported in table \ref{tb}. We report our model's defense accuracy, substitute model's accuracy, substitute model's accuracy under attack and our defense model's accuracy under attack sequentially in the table. We use full-precision activation at the source in order to precisely approximate the decision boundaries of the target model. It is observed that our defense's accuracy degrades to some extent when attacked using a VGG16 substitute model. However, it is much better than the substitue DNN model under the same attack. Using a 32bit VGG-16 model at the source could generate attack that would give as low as 49.18 \% accuracy for binary activation. An obvious reason would be using a VGG16 model would generate very strong adversarial examples due to the model's increased capacity and full- precision activation. However, for Alexnet substitute model we could get much higher accuracy than the white box counter part. Our defense could achieve 82.59, 82.11 and 79.74 \% accuracy for 3,2,1 bits respectively. The reported accuracies are much higher than the white box attack and also higher than the accuracy achieved attacking the substitute model.Our proposed DQA defense preforms reasonably well against both substitute attack model and ZOO model to further establish our hypothesis. As suppressing the noise using lower bit-width activation and adding more parameters in the adversarial training process certainly makes the DNN more robust against both white box and black box attack.

\section{Summary}
In this work, we first propose to employ quantized activation function in DNN to defend adversarial examples. Further, we utilize the benefit of using adversarial training and quantized activation function even more by introducing dynamic quantization. Dynamic quantization pushes the state-of-the-art defense accuracy of both MNIST and CIFAR 10 further under strong attacks. Our hypothetical analysis is proven empirically through a variety of experiments showing both fixed and dynamic quantization can provide strong resistance to adversarial attack. Thus network compression and model robustness can be achieved at the same time which may lead to further compact robust neural networks research.

\bibliographystyle{unsrt}
\bibliography{reference}

\end{document}